\documentclass[letterpaper]{article} 
\usepackage{aaai24}  
\usepackage{times}  
\usepackage{helvet}  
\usepackage{courier}  
\usepackage[hyphens]{url}  
\usepackage{graphicx} 
\usepackage{natbib}  
\usepackage{caption} 
\usepackage{algorithm}
\usepackage{algorithmic}

\usepackage{newfloat}
\usepackage{listings}

\def\ie{\textit{i.e.}}

\DeclareCaptionStyle{ruled}{labelfont=normalfont,labelsep=colon,strut=off} 
\floatstyle{ruled}
\newfloat{listing}{tb}{lst}{}
\floatname{listing}{Listing}
\title{LaneGraph2Seq: Lane Topology Extraction with Language Model via Vertex-Edge Encoding and Connectivity Enhancement}
\author{
    Renyuan Peng\textsuperscript{\rm 1},
    Xinyue Cai\textsuperscript{\rm 2},
    Hang Xu\textsuperscript{\rm 2}\equalcontrib,
    Jiachen Lu\textsuperscript{\rm 1},
    Feng Wen\textsuperscript{\rm 2},
    Wei Zhang\textsuperscript{\rm 2},
    Li Zhang\textsuperscript{\rm 1}\equalcontrib
}
\affiliations{
    \textsuperscript{\rm 1}School of Data Science, Fudan University \quad \\
    \textsuperscript{\rm 2}Huawei Noah’s Ark Lab \\
    chromexbjxh@gmail.com, lizhangfd@fudan.edu.cn
    
}

\usepackage{bibentry}

\begin{document}

\maketitle

\begin{abstract}
Understanding road structures is crucial for autonomous driving. 
Intricate road structures are often depicted using lane graphs, which include centerline curves and connections forming a Directed Acyclic Graph (DAG). 
Accurate extraction of lane graphs relies on precisely estimating vertex and edge information within the DAG.
Recent research highlights Transformer-based language models' impressive sequence prediction abilities, making them effective for learning graph representations when graph data are encoded as sequences. 
However, existing studies focus mainly on modeling vertices explicitly, leaving edge information simply embedded in the network. 
Consequently, these approaches fall short in the task of lane graph extraction.
To address this, we introduce \textbf{\textit{LaneGraph2Seq}}, a novel approach for lane graph extraction. 
It leverages a language model with vertex-edge encoding and connectivity enhancement. 
Our serialization strategy includes a vertex-centric depth-first traversal and a concise edge-based partition sequence. 
Additionally, we use classifier-free guidance combined with nucleus sampling to improve lane connectivity.
We validate our method on prominent datasets, nuScenes and Argoverse 2, showcasing consistent and compelling results. 
Our LaneGraph2Seq approach demonstrates superior performance compared to state-of-the-art techniques in lane graph extraction.
Code is available at \url{https://github.com/fudan-zvg/RoadNet}

\end{abstract}
\section{Introduction}
\begin{figure}[htb]
\vspace{-2mm}
    \centering
    \includegraphics[width=\linewidth]{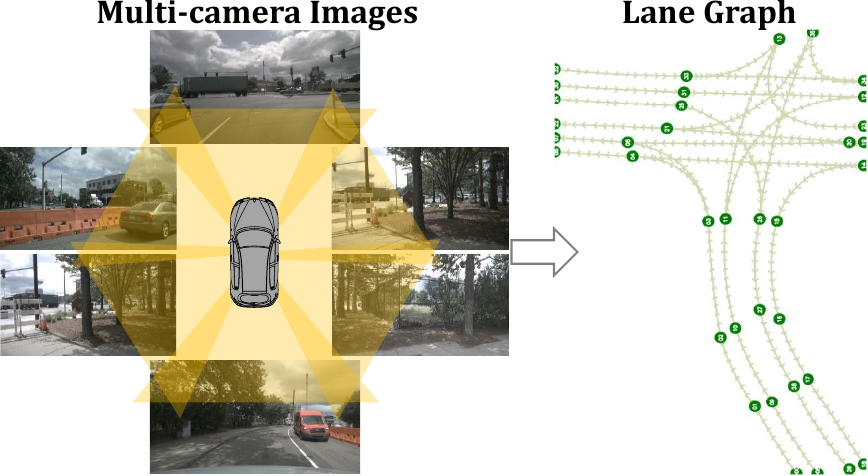}
    \vspace{-4mm}
    \caption{The extraction of a lane graph, which captures the centerline curves and their intricate connectivity relationships on the road, is crucial for the perception system of autonomous driving. Our proposed \textit{LaneGraph2Seq} framework adeptly addresses this challenge by utilizing multi-camera images as input.
    }
    \label{fig:intro}
    \vspace{-3mm}
\end{figure}
Road structure understanding serves as a foundational pillar that empowers autonomous vehicles to navigate safely, efficiently, and intelligently within complex real-world scenarios~\cite{cui2019multimodal, chen2020learning, hong2019rules, espinoza2022deep}.
A road structure is suitable to be represented by a Directed Acyclic Graph(DAG) due to its characteristics of one-way traffic flow, acyclic layout, complex lane topologies, and meaningful spatial relationships.
The attributes of a DAG align with the inherent features of roads, making it a potent and intuitive representation for modeling and analyzing road structures. 
Hence, the complex road network is frequently represented as a directed acyclic lane graph, which includes the \textbf{\textit{centerline curves}} and their \textbf{\textit{connection relationships}}~\cite{can2021structured, can2022topology, buchner2023learning}.
Nevertheless, extracting a lane graph using onboard sensors is inevitable to be a challenging task due to the intricate nature of the road structure.

The prevailing frameworks for lane graph extraction often draw inspiration from models like STSU~\cite{can2021structured} and TPLR~\cite{can2022topology}. 
These frameworks model centerline curves as vertices and represent the interconnections between centerlines as edges. 
In the process of lane graph extraction, 
they employed a DETR-like module for detecting the centerline's shape and position.
Subsequently, an MLP is employed to approximate the connectivity between any pair of centerlines.
This two-stage framework predicts the vertices and edges of the lane graph with a relatively independent approach. 
While utilizing visual features, this method falls short in terms of overall graph feature learning.

Graph Neural Network(GNN) and its variants are widely adopted to aggregate features of vertices and extract information from graph data~\cite{velickovic2017graph,hamilton2017inductive,kipf2016semi}. 
Recently, Graphormer~\cite{ying2106transformers} has demonstrated that by adeptly encoding the structural information of a graph into the model, the standard Transformer architecture can surpass GNNs and  achieve remarkable outcomes across a wide spectrum of graph representation learning tasks.
However, Graphormer solely focuses on explicitly encoding the vertices of the graph into sequences. 
In the context of lane graph extraction, the task necessitates the prediction of both vertices and edges. 
This crucial distinction renders existing Transformer-based graph representation learning methods, such as Graphormer, unsuitable for the specific demands of lane graph extraction.

To enhance the acquisition of graph features and facilitate lane graph extraction, we present \textbf{\textit{LaneGraph2Seq}}—a novel approach utilizing a Transformer-based language model with vertex-edge encoding and connectivity enhancement.
Precisely, our approach employs the language model with Transformer architecture to grasp the intricacies of the lane graph. 
By representing the graph's vertices and edges as a sequence of discrete tokens, we encapsulate the complete information of a lane graph within it.
Consequently, this challenge, graph prediction, is transmuted into a sequence prediction task. 
Leveraging bird's-eye-view(BEV) encoder and language model with Transformer architecture, we achieve the dual objectives of acquiring graph features from onboard sensors and predicting graph sequences.
Furthermore, to mitigate the occurrence of premature termination or continuous loops in sequence prediction, enhancing the diversity of language model prediction samples proves to be a effective strategy.
Hence, we employ classifier-free guidance~\cite{ho2022classifier} and nucleus sampling~\cite{holtzman2019curious} to augment the connectivity conditions and increase the sampling diversity. 
This mechanism enables the network to strike a balance between accuracy and completeness in edge prediction.

The \textbf{\textit{contributions}} of this work are summarized as follows: 
\textbf{(i)} We introduce an innovative framework for lane graph extraction, known as \textbf{\textit{LaneGraph2Seq}}, which casts the task as a sequence-to-sequence prediction challenge. 
Specifically, we present a vertex-edge encoding approach for sequence construction and employ a language model to simultaneously analyze visual features and extract graphical attributes.
\textbf{(ii)}
During the inference stage, we employ a method that combines classifier-free guidance with nucleus sampling to enhance the diversity of sampling. This approach contributes to improved accuracy in edge prediction and a reduction in the false negative rate by enhancing lane graph connectivity conditions.
\textbf{(iii)} Extensive experiments conducted on two large-scale datasets (nuScenes~\cite{caesar2020nuscenes} and Argoverse 2~\cite{wilson2023argoverse}) demonstrate that our approach attains state-of-the-art performance in lane graph extraction (Figure~\ref{fig:intro}).

\section{Related Work}
\paragraph{Bird’s-Eye-View (BEV) Semantics Learning}
Recently, there's been growing interest in transforming monocular or multi-view images from ego car cameras into bird's-eye-view (BEV) representations.
To learn BEV feature from onboard cameras, OFT~\cite{roddick2018orthographic}, LSS~\cite{philion2020lift} and FIERY~\cite{hu2021fiery} predict 3D spatial features from front view images through real or predicted depth information, and then obtain BEV features from 3D spatial features.
\cite{saha2022translating, roddick2020predicting, li2022bevformer, lu2022learning} leverage a Transformer to implicitly learn 3D spatial information and get features in the bird's eye view.
To capture the lane graph represented by BEV, we employ the LSS technique works~\cite{philion2020lift, huang2021bevdet} to extract BEV features from onboard cameras.

\paragraph{Lane Graph Extraction}
Previously, extensive research has been conducted on the online construction of High-Definition (HD) Maps~\cite{li2022hdmapnet, liu2023vectormapnet}.
Recently, to extract the lane graph online, STSU~\cite{can2021structured} first identifies centerlines using a Transformer-based approach on image features.  Subsequently, it forecasts centerline associations using an MLP layer, culminating in a final merging step to construct the lane graph.
Based on STSU, TPLR~\cite{can2022topology} introduces a minimal cycle to eliminate ambiguity in its connectivity representation.
Furthermore, LaneGAP~\cite{liao2023lane} adopts an innovative pathwise approach to modeling the lane graph, effectively preserving the lane's continuity.
 
While the previously mentioned methods concentrate on local visual attributes, they overlook the lane graph's inherent features. 
In contrast, our approach involves encoding the lane graph into sequences and leveraging a language model for predictions. 
This methodology not only facilitates the acquisition of visual features but also comprehends the graph's distinctive characteristics.

\paragraph{Language Modeling}
While initially conceived for natural languages, language modeling has demonstrated its aptitude in modeling a range of sequential data tasks~\cite{sutskever2014sequence, raffel2020exploring, he2016deep}.
Motivated by these achievements in natural language processing, contemporary efforts within the realm of computer vision and vision-language have also begun to delve into the utilization of language modeling for various tasks \cite{chen2021pix2seq, chen2022unified, wang2022ofa, liu2023polyformer}.

These instances of success inspire us that language models excel not only in comprehending linguistic logic but also exhibit proficiency in visual localization.
This demonstrates that employing sequence prediction methods enables the language model to effectively predict both road topology (centerline connection) and the shape of centerline curves.

\paragraph{Graph Represent Learning}
Graph Neural Networks (GNNs)~\cite{scarselli2008graph, kipf2016semi, hamilton2017inductive, velickovic2017graph} play a pivotal role in aggregating information and extracting insights from graph-structured data.
This methodology has also been employed in recent endeavors to perceive autonomous driving scenarios~\cite{weng2021ptp, weng2020gnn3dmot}. 
Notably, LaneGCN~\cite{liang2020learning} constructs a lane graph from HD maps, while TopoNet~\cite{li2023topology} employs a GNN model to accomplish road network extraction and detect traffic elements. 
Recent research~\cite{ying2106transformers} has further demonstrated that the Transformer architecture can surpass GNNs in a broad spectrum of graph-level prediction tasks when the graph information is pertinently embedded.

Inspired by these findings, we adopt a vertex-edge encoding method to serialize the lane graph and subsequently extract lane graph using a Transformer structure.

\section{Method}
\begin{figure}[tb]
    \centering
    \includegraphics[width=\linewidth]{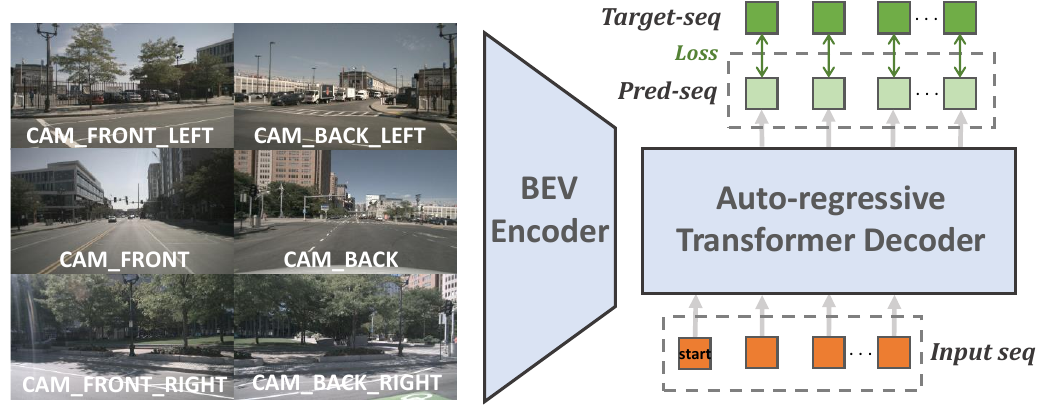}
    \caption{Our LaneGraph2Seq employs a BEV-encoder to transition features from the front view image to the bird's-eye view plane. Subsequently, a Transformer decoder generates tokens of the target sequence in sequence, guided by prior tokens and the encoded BEV feature.
    }
    \vspace{-3mm}
    \label{fig:model}
\end{figure}
\begin{figure}[tb]
    \centering
    \includegraphics[width=\linewidth]{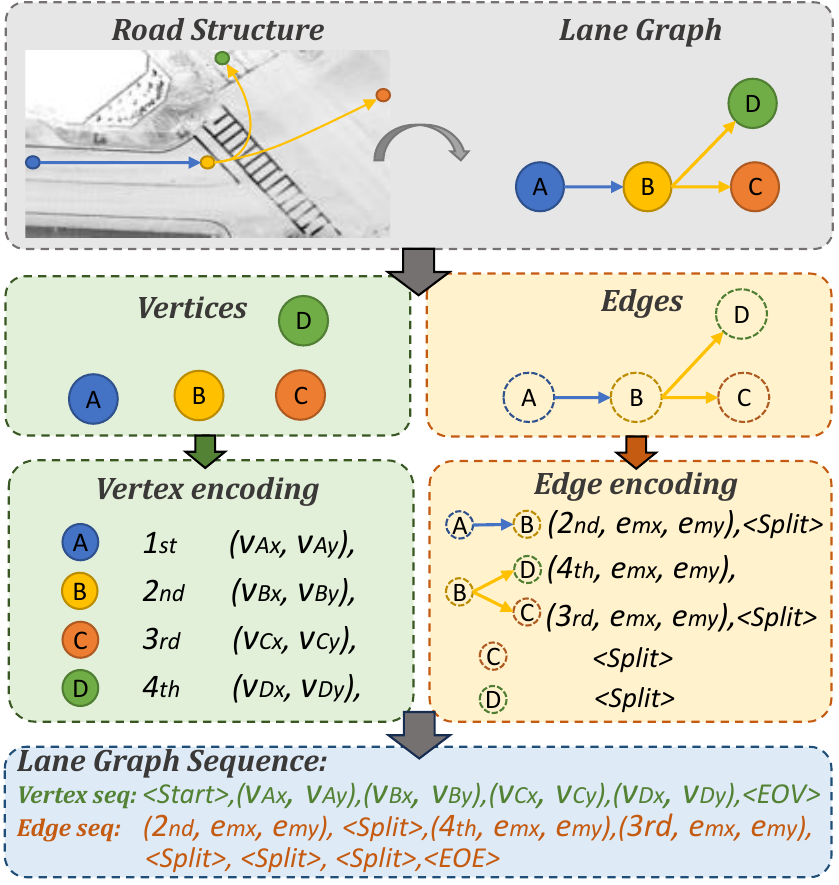}
    \vspace{-4mm}
    \caption{This depiction outlines the procedure for constructing a sequence that represents the actual road structure. 
    The upper part illustrates the abstraction of the real road into a Directed Acyclic Graph (DAG), while the middle section showcases the detailed process of encoding vertices and edges. 
    The lower part exhibits the resulting vertex sequence and edge sequence after vertex-edge encoding, which is then combined to form the comprehensive lane graph sequence.
    }
    \label{fig:seq}
    \vspace{-3mm}
\end{figure}

\paragraph{Lane Graph Representation}
The lane graph comprises centerlines and their associated junction points~\cite{can2021structured, can2022topology}.
Because traffic flows in a single direction on each lane, the graph can be represented as a Directed Acyclic Graph (DAG), \ie, $G = (V, E)$ where the vertex set $V$ is the set of all junction points and the edge set $E$ is the set of all centerlines.
Each vertex $v = (v_x, v_y) \in V$ holds the coordinates of junction points.
Each edge $e=(e_{s}, e_{m}, e_{t})\in E$ is characterized by three Bezier control points: the source point, the midpoint, and the target point.
The shape of the curve can be determined by the coordinates of the three control points in this specific order.

\paragraph{Overall Structure of the Sequence}
As shown in Figure~\ref{fig:seq}, the sequence of a lane graph comprises two components: the \textit{\textbf{vertex sequence}} and the \textit{\textbf{edge sequence}}.

The vertex sequence consists of vertices coordinates arranged in a specific order.
Since lane graphs encompass varying vertex quantities, the resulting sequences will exhibit distinct lengths.  
In order to demarcate the conclusion of a vertex sequence, we introduce an \texttt{<EOV>} (End of Vertex Sequence) token.

As mentioned before, an edge can be represented as a set of the source vertex, the target vertex, and the Bezier middle control point.
For each vertex in DAG, We denote the target vertex using the sequential index of its child nodes within the vertex sequence.
The source vertex of each edge is determined by arranging the edge in the order of its source vertex in the vertex sequence.

So with three parameters $( [Index(child node), e_{mx}, e_{my}] )$ and the position of this triple, the curve shape and direction of an edge can be determined.
Due to the variable nature of the out-degree for each vertex in the DAG, the number of edges extended by each vertex may vary. Consequently, we employ a \texttt{<Split>} token to ascertain the end of the edge subsequence corresponding to a given vertex as the parent node.
And at the end of the entire edge sequence, we set the \texttt{<EOE>}(End of Edge Sequence) token to determine the termination.

Besides, we use \texttt{<Start>} token to indicate the beginning of the whole sequence. 
The \texttt{<N/A>} token is utilized to separately pad the vertex sequence and edge sequence, which guarantees that both of them maintain a consistent length.

\paragraph{Serialization}

\begin{figure}[tb]
    \centering
    \includegraphics[width=\linewidth]{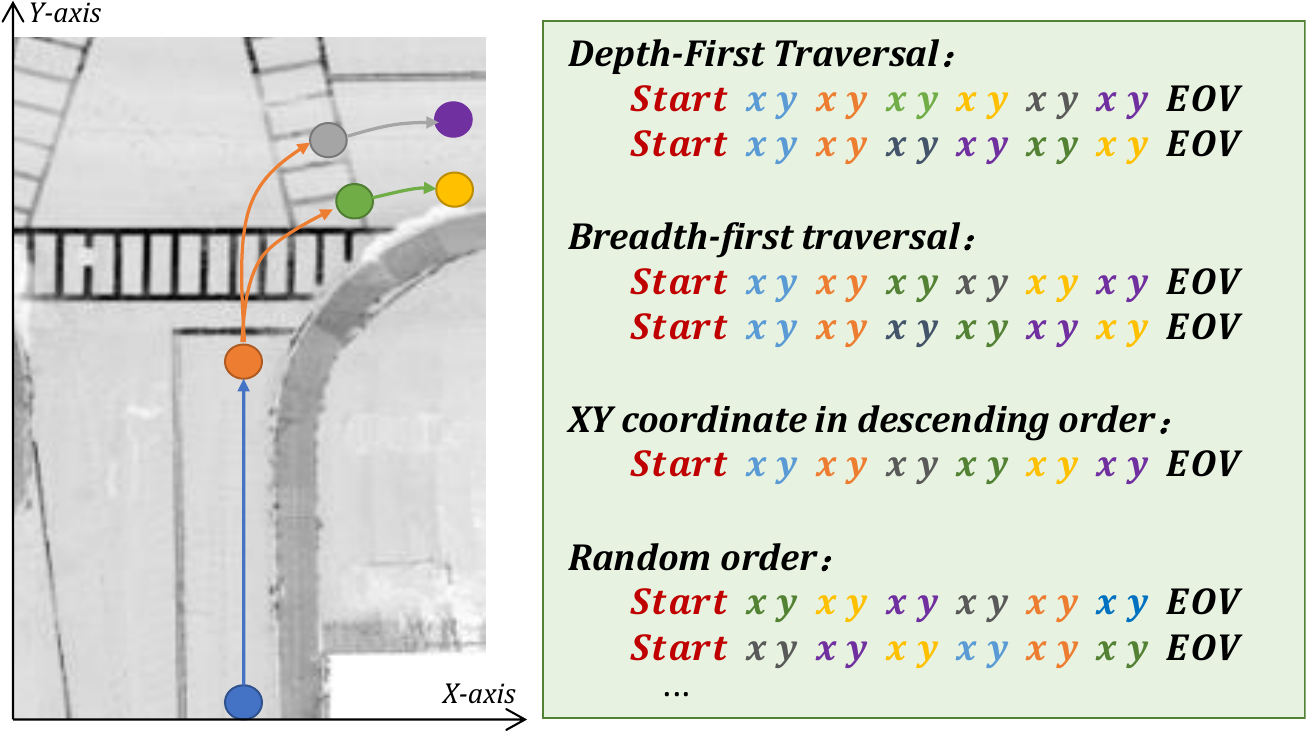}
    \vspace{-4mm}
    \caption{
    Examples of serialization order.
    \textit{Right} provides possible example vertex sequences of different sorting methods. 
    \textit{Start} denotes the \texttt{<start>} token, while \textit{EOV} signifies the \texttt{<EOV>} token. 
    The differently colored \textit{x y} entries indicate the x and y coordinate values of points corresponding to the colors on the \textit{left}. 
    As edge order aligns with point order, we exclusively present the vertex sequence.
    }
    \label{fig:order}
    \vspace{-2mm}
\end{figure}
While the order of edges is linked to the sequence of vertices, the exact positioning of the vertices remains uncertain. 
Hence, serializing the vertices into a deterministic sequence becomes essential for constructing a lane graph.
In contrast to detection and segmentation tasks~\cite{chen2021pix2seq, liu2023polyformer}, there are reasons to believe that the traversal order of the lane graph impacts the eventual performance.
We explored several deterministic ordering strategies, including depth-first traversal, breadth-first traversal, sorting by coordinate value magnitude, and random sorting. 
Figure~\ref{fig:order} depicts the sequence construction process using distinct ordering strategies. 
We assume that utilizing a depth-first traversal order is better suited for enabling the auto-regressive Transformer network to autonomously explore and reveal the entirety of the lane graph.

\paragraph{Discretization}
While the indices and tokens with special significance (\texttt{<Start>}, \texttt{<EOV>}, \texttt{<EOE>}, \texttt{<Split>}, \texttt{<N/A>}) are represented as discrete tokens, the coordinates of the vertices and Bezier middle control points are not treated discretely.
Following ~\cite{chen2021pix2seq}, We discretize the continuous coordinate values by partitioning the image into varying numbers of bins. 
The greater the number of bins, the smaller the quantization error. 
We use a shared fixed vocabulary for all tokens.
It's worth noting that, for the network to learn distinct token classes, these classes are distributed across varying ranges within the vocabulary. 
For instance, if vertex coordinates fall within the range of $[1, num\_bins]$, and an offset constant would be added to the Bezier control point coordinates, shifting their values into the range of $[num\_bins, num\_bins * 2]$.
Furthermore, tokens with special significance fall outside of these ranges.

\paragraph{Architecture}

Just like in various other downstream tasks within autonomous driving, our approach involves initially utilizing a BEV-encoder to transform the features from the front view images onto the bird's-eye view plane. Subsequently, we employ a decoder like~\cite{chen2021pix2seq}, which generates tokens of the target sequence sequentially based on preceding tokens and the encoded BEV feature. The overall architecture is illustrated in the Figure~\ref{fig:model}.

\paragraph{Objective}
Given the ground-truth sequence denoted as $y$ with a length of $L$, and the predicted sequence denoted as $\hat{y}$, the model's objective is to minimize the maximum likelihood loss, which can be expressed as:
\begin{equation}
\max \sum_{i=1}^{L} w_i \log{P(\hat{y_i}|y{<i}, \mathcal{F})}
\end{equation}
Here, $\mathcal{F}$ is the BEV feature, $y_i$ represents the $i^{th}$ token of $y$, $y_{<i}$ signifies all tokens preceding $y_i$, and $w_i$ stands for the class weight.

In practical implementation, to prevent premature termination predictions, we reduce the weights of \texttt{<EOV>}, \texttt{<Split>}, and \texttt{<EOE>} tokens accordingly.
The \texttt{<N/A>} token, employed for padding, does not contribute to the backpropagation of loss.

\paragraph{Inference with Connectivity Enhancement}
The \texttt{<EOS>} and \texttt{<EOE>} tokens empower the model to determine when to conclude the generation of the vertex sequence and the edge sequence.  
Similarly, the \texttt{<Split>} token enables the model to ascertain when to halt the generation of a sub edge sequence pertaining to a vertex. 
In practice we find that the model tends to finish without predicting all edges.

Classifier-free guidance entails guiding an unconditioned sample towards a conditioned counterpart~\cite{gafni2022make}. 
This approach directs a model's generation process without the need for explicit classifiers or labels. 
It finds application in various domains such as text generation, image synthesis, and sequence prediction, offering increased flexibility and creativity in outputs while avoiding limitations of traditional classification methods.

The vertex sequence encompasses the coordinates of all junction points, encapsulating the lane graph's connectivity information and providing a strong prior for generating the edge sequence.
So we take the vertex sequence as condition to generate the edge sequence.
Continuing the approach from previous research~\cite{gafni2022make}, in the inference phase, we generate two simultaneous token streams: a conditioned token stream dependent on the vertex sequence, and an unconditioned token stream linked to a mask token.
For the Transformer decoder, we apply classifier-free guidance on logit scores:

\begin{equation}
logits_{cond} = T(t_e | (t_v, \mathcal{F}))
\end{equation}
\begin{equation}
logits_{uncond} = T(t_e | (<mask>, \mathcal{F}))
\end{equation}
\begin{equation}
logits_{cf} = logits_{uncond} + \alpha_c (logits_{cond} - logits_{uncond})
\end{equation}

where $T$ represents the Transformer decoder, $t_e$ and $t_v$ represent the token of edge sequence and vertex sequence respectively, $\mathcal{F}$ stands for the BEV feature, and $\alpha_c$ is the scale factor.

Through the implementation of classifier-free guidance and utilizing the vertex sequence as a condition, the model increases the diversity of samples on the basis of enhanced connectivity during edge sequence prediction. 
This results in more comprehensive and accurate edge predictions.

\begin{figure*}[tb]
    \centering
    \includegraphics[width=\linewidth]{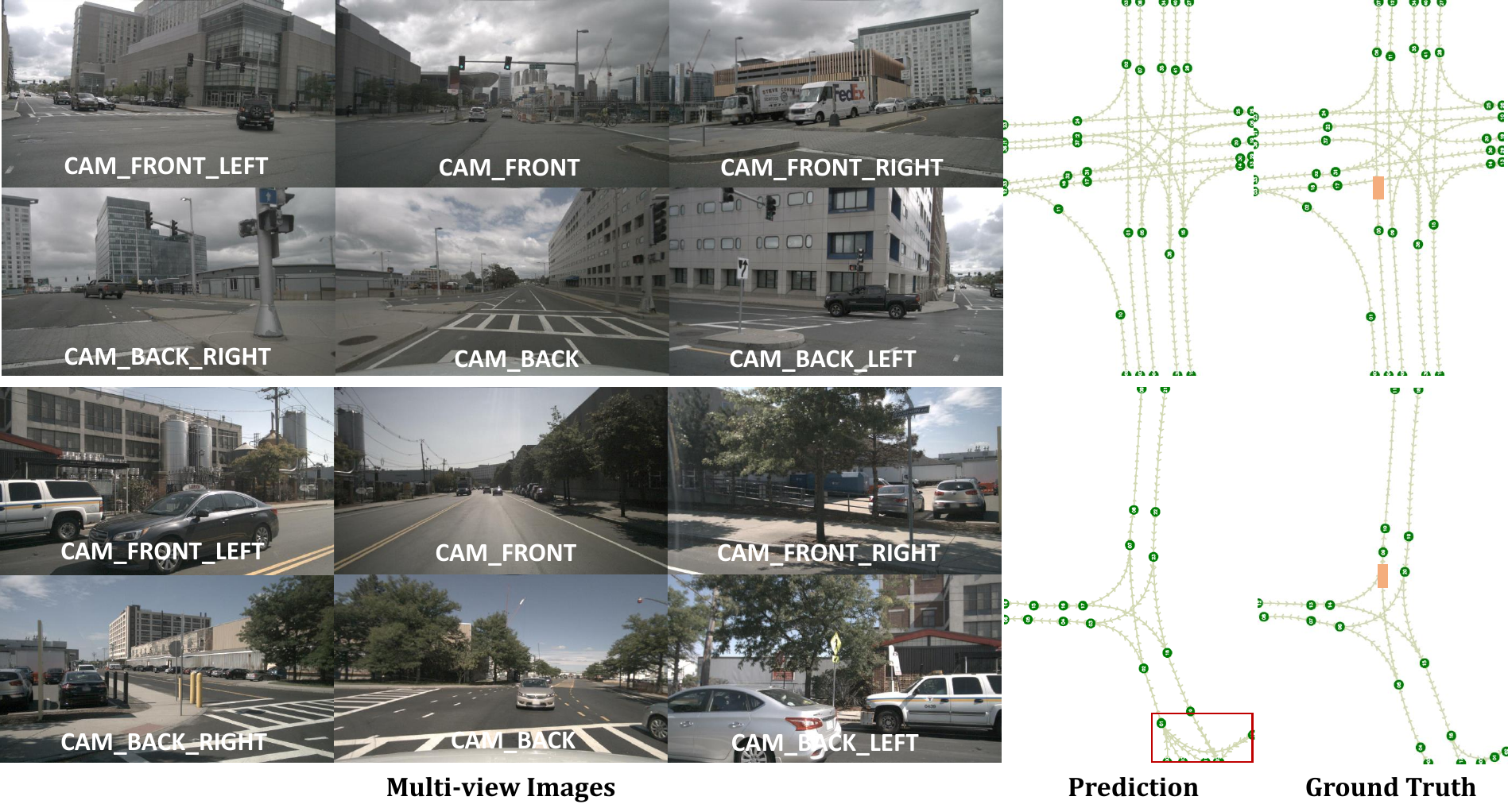}
    \vspace{-4mm}
    \caption{Our qualitative results on nuScenes~\cite{caesar2020nuscenes} validation set. Evidently, our approach demonstrates an impressive ability to attain highly accurate predictions, with only a slight error in the red-boxed section.
    }
    \label{fig:nus_vis}
    \vspace{-2mm}
\end{figure*}
\begin{figure*}[tb]
    \centering
    \includegraphics[width=\linewidth]{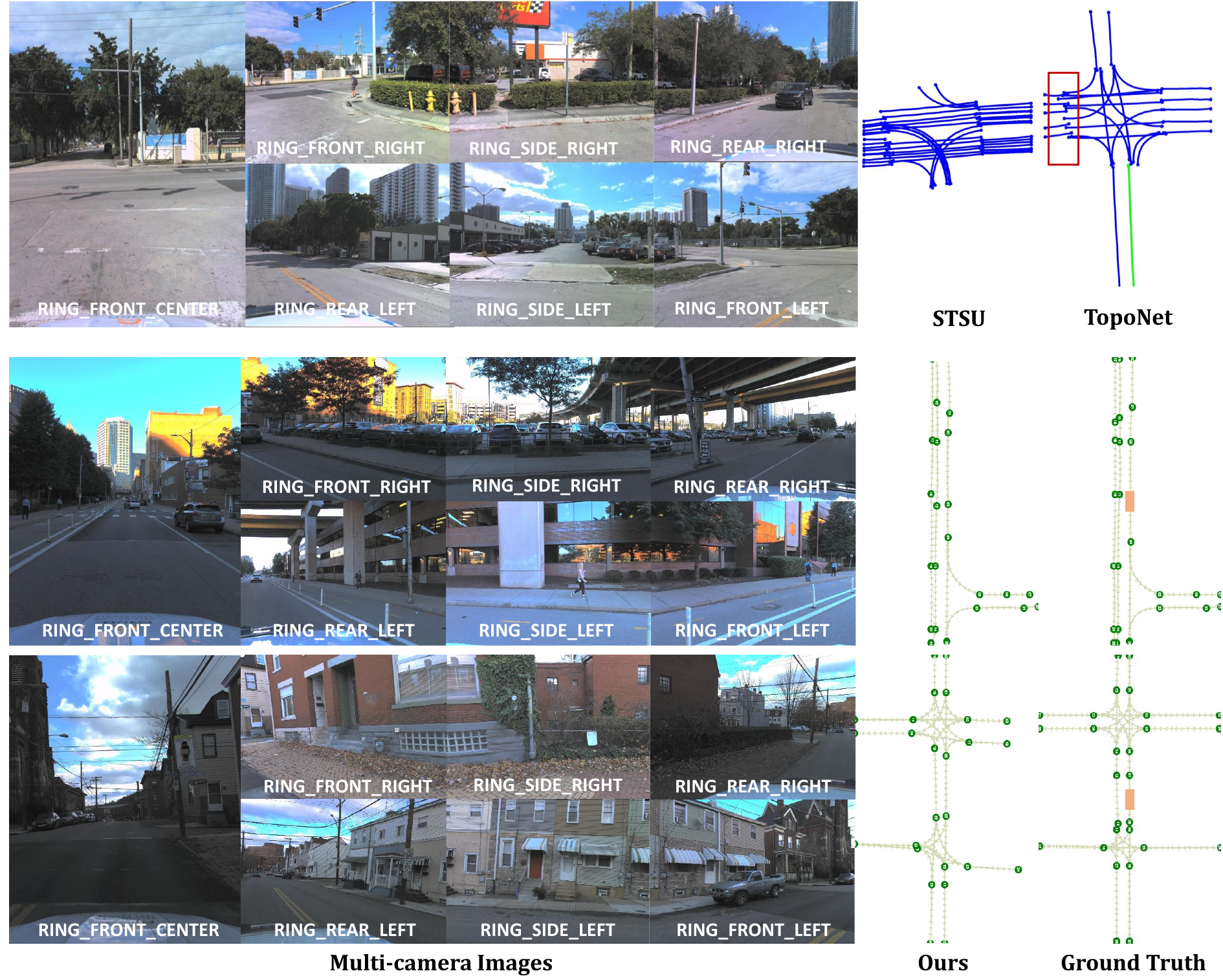}
    \vspace{-5mm}
    \caption{Qualitative outcomes on the Argoverse 2~\cite{wilson2023argoverse} validation set. 
    The first row~\cite{li2023topology} shows the visualization result of STSU~\cite{can2021structured} and TopoNet~\cite{li2023topology}.
    The results indicate that STSU inaccurately predicts the road layout, and TopoNet displays minor deficiencies in the vicinity of junction points.
    The results show that our approach effectively predicts the overall layout of the road structure with accuracy, especially with continuous and clear junction points.
    }
    \label{fig:av2_vis}
    \vspace{-3mm}
\end{figure*}

\paragraph{Sequence to Lane Graph}
Initially, we identify the first \texttt{<EOV>} and \texttt{<EOE>} tokens in the sequence, extracting the vertex sequence and edge sequence without padding. 
By detecting \texttt{<Split>} tokens, we divide the edge sequence into subsequences. Each subsequence signifies edges, with the node located at the corresponding position serving as the parent node.
Hence, with all the vertices and edges in place, the lane graph is reconstructed through the sequence.

\section{Experiments}

\subsection{Dataset}
We conducted benchmarking on two challenging datasets: 

\noindent\textbf{nuScenes}~\cite{caesar2020nuscenes}The comprises 1000 sequences, officially divided into train, validation, and test sets with 700, 150, and 150 scenes, respectively. Each sequence is captured at a 2Hz frame rate, offering RGB images from 6 surrounding cameras, covering a 360-degree horizontal field of view around the ego-vehicle.

\noindent\textbf{Argoverse 2}~\cite{wilson2023argoverse} is even more extensive, featuring 1000 sequences collected from 6 different cities. 
It is randomly partitioned into the train, validation, and test splits, with ratios of 700, 150, and 150 sequences respectively. 
Argoverse 2 provides images from 7 surrounding cameras and offers a significantly larger volume of data, with four times the amount available in the nuScenes.

Both datasets offer lane graph information represented as lane centerlines. 
In our online lane graph construction scenario, the target BEV range extends from -48 to 48m along the X-axis and -32 to 32m along the Y-axis, with points sampled at intervals of 0.5m. 
We conduct training and evaluation using the specified official training and validation sets for both datasets with all available surrounding-view images.

\subsection{Implementation Details}
Our BEV encoder, following the approach of LSS~\cite{philion2020lift}, employs either the ResNet50~\cite{he2016deep} or VovNetV2~\cite{lee2020centermask} backbone to transform input RGB images into BEV features.
For improved initialization, we pretrain the BEV encoder on the task of centerline segmentation.

For sequence construction, we pad the vertex sequence to 200 and the edge sequence to 400, resulting in a total lane sequence length of 600.
The Transformer decoder layers are set to 6.
We trained the network for 300 epochs on 8 NVIDIA V100 GPUs using the AdamW optimizer, with an initial learning rate of $2 \times 10^{-4}$ and a batch size of $2 \times 8$.
During the training process, we apply random flip, random rotation, and random scaling on BEV feature similar to~\cite{huang2021bevdet,lu2022learning}.
In the inference process, we configure the $\alpha_c$ parameter of the classifier-free guidance to 4.

\subsection{Metrics}
To assess the accuracy of lane graph extraction, we employed the identical evaluation metrics as those utilized in STSU~\cite{can2021structured}:

\noindent\textbf{Precision-Recall} (M-P, M-R, M-F ratio)
measures how closely the estimated centerlines fit the matched ground truth (GT) centerlines and how accurately the subgraph is captured. This metric focuses on matching estimated centerlines to GT ones, without penalizing unmatched true centerlines.

\noindent\textbf{Detection ratio} (Detect)
quantifies the portion of distinct ground truth centerlines matched by at least one estimated line, addressing the precision-recall metric's limitation by considering missed centerlines. High precision-recall and low detection ratio suggest proximity between estimated and matched GT lines, yet numerous GT centerlines go undetected.

\noindent\textbf{Connectivity} (C-P, C-R, C-F ratio)
evaluates the association and connectivity between the estimated centerlines using a precision-recall-based approach. 
This metric focuses on the edge of the lane graph and measures whether the estimated centerlines have a similar connection relationship with the ground truth.

\subsection{Comparison with the State-of-the-Art}
\begin{table*}[tb]
  \centering
  \setlength{\tabcolsep}{4mm}
    \begin{tabular}{c||c|ccc|c|ccc}
    \hline
    
    \hline
    Methods & Dataset & M-P & M-R & M-F & Detect & C-P & C-R & C-F\\
    \hline

    PINET~\cite{ko2021key} & nuScenes & 54.1 & 45.6 & 49.5 & 19.2 & - & - & -\\
    Poly~\cite{acuna2018efficient}& nuScenes  & 54.7 & 51.2 & 52.9 & 40.5 & 58.4 & 16.3 & 25.5\\
    STSU~\cite{can2021structured}& nuScenes  & 60.7 & 54.7 & 57.5 & 60.6 & 60.5 & 52.2 & 56.0\\
    TPLR~\cite{can2022topology}& nuScenes  & - & - & 58.2 & 60.2 & - & - & 55.3\\
    \hline
    LaneGraph2Seq& nuScenes  & 64.6 & 63.7 & 64.1 & 64.5 &  69.4 & 58.0 & 63.2\\

    LaneGraph2Seq\dag & nuScenes & \textbf{68.1}& \textbf{79.3} & \textbf{73.3} & \textbf{68.7} & \textbf{75.2}& \textbf{61.4}& \textbf{67.6}\\
    \hline
    \hline
    LaneGraph2Seq & AV2 & 62.6 & 60.9 & 61.7 & 62.6 &  60.1 & 57.3 & 60.1\\
    LaneGraph2Seq\dag & AV2 & \textbf{65.6} & \textbf{70.7} & \textbf{68.0} & \textbf{65.9} &  \textbf{66.5} & \textbf{59.8} & \textbf{63.9}\\

    \hline
    \end{tabular}
    \vspace{-3mm}
    \caption{Comparison of state of the art and our method on nuScenes dataset and Argoverse 2 dataset. 
    ResNet-50~\cite{he2016deep} is applied as image backbone by default.
    ``\dag" use VoVNetV2~\cite{lee2020centermask} pretrained on extra data as backbone.
    \texttt{M-P}, \texttt{M-R}, \texttt{M-F} stand for mean precision/recall/F1-score.
    \texttt{Detect} stands for Detection ratio metrics.
    \texttt{C-P}, \texttt{C-R}, \texttt{C-F} stand for connectivity precision/recall/F1-score.
    }
  \label{tab:nus-stsu}
  \vspace{-2mm}
\end{table*}

We compare our proposed model with previous state-of-the-art methods for lane graph extraction on the nuScenes dataset.
Detailed outcomes of this comparison are presented in Table~\ref{tab:nus-stsu}. Impressively, our model outperforms all prior including STSU~\cite{can2021structured} and TPLR~\cite{can2022topology} approaches across every evaluation metric. 
Notably, Precision-Recall metrics show an improved match between estimated centerlines and ground truth. 
This notable improvement results from innovative language modeling techniques, including vertex-edge encoding and connectivity enhancement.

\begin{table}[tb]
  \centering
  \setlength{\tabcolsep}{5mm}
    \begin{tabular}{c||c|c|c}
    \hline
    
    \hline
    Order & M-F & Detect & C-F\\
    
    \hline
    $DFS$ & \textbf{73.3} & \textbf{68.7} & \textbf{67.6}\\
    $BFS$ & 72.1& 67.2 & 67.2\\
   $Coord_{xy}$ & 69.8 & 61.7 & 61.2\\
    $Random$ & 70.2 & 61.3 & 62.1\\
    \hline

    \hline
    \end{tabular}
    \vspace{-1mm}
    \caption{Ablation study on serialization order with six cameras as input on nuScenes. 
    VoVNetV2 is applied as image backbone by default.
    \texttt{$DFS$} and \texttt{$BFS$} denote depth-first traversal and breadth-first traversal, while \texttt{$Coord_{xy}$} indicates sorting vertices based on their xy coordinate values in ascending order, and \texttt{$Random$} is a random arrangement.
    }
  \label{tab:order}
\end{table}

\begin{table}[tb]
  \centering
  \setlength{\tabcolsep}{5mm}
    \begin{tabular}{c||c|c|c}
    \hline
    
    \hline
    Layers & M-F & Detect & C-F\\
    \hline
    3 & 70.8 & 63.5 & 63.6\\
    4 & 72.0& 65.2 & 64.2\\
    6 & 73.3 & 68.7 & 67.6\\
    8 & \textbf{74.0} & \textbf{69.3} & \textbf{69.1}\\
    \hline

    \hline
    \end{tabular}
    \vspace{-1mm}
    \caption{Ablation study on number of Transformer layers with six cameras as input on nuScenes dataset. 
    VoVNetV2 is applied as image backbone by default. 
    }
  \label{tab:layers}
  \vspace{-3mm}
\end{table}

\subsection{Ablation Studies}
We perform a comprehensive set of ablation studies on the validation set of nuScenes~\cite{caesar2020nuscenes}.
All the results presented in this section were obtained utilizing six surrounding-view images as input.

\begin{figure}[tb]
\vspace{-2mm}
    \centering
    \includegraphics[width=\linewidth]{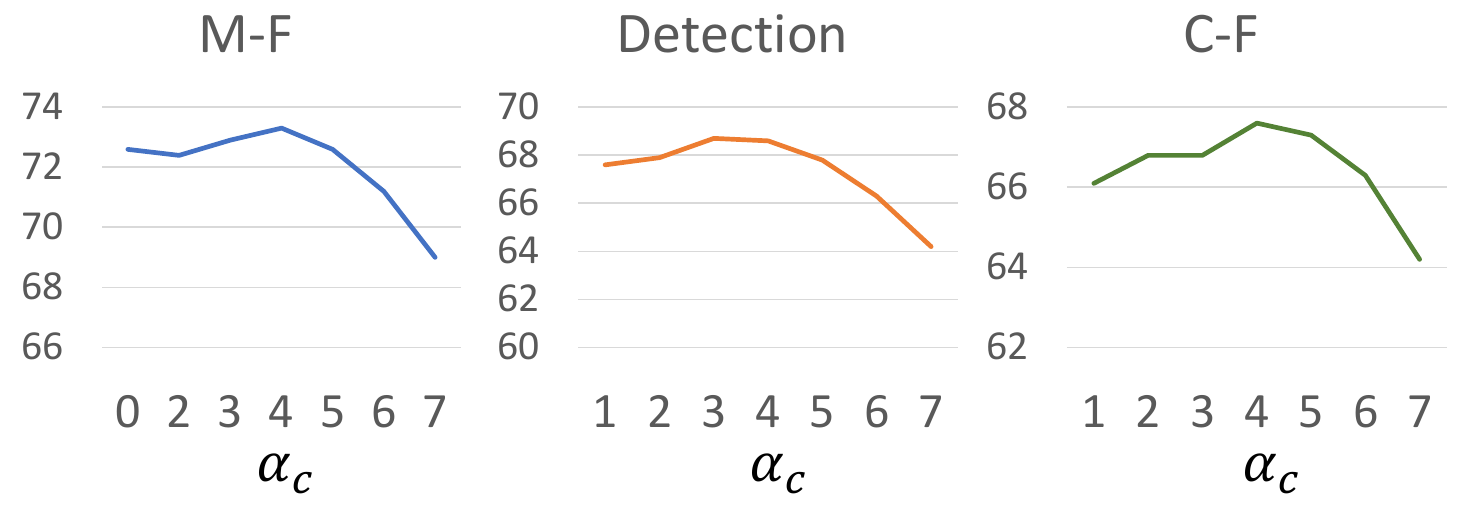}
    \vspace{-3mm}
    \caption{Altering parameter $\alpha_c$ in classifier-free guidance yields distinct \texttt{M-F}, \texttt{Detect}, and \texttt{C-F} outcomes.
    When $\alpha_c$ is set to 1, it equates to inference without connectivity enhancement.
    }
    \label{fig:cfg}
    \vspace{-3mm}
\end{figure}

\paragraph{Connectivity Enhancement}
Building upon the insights from the ablation study on the nuScenes dataset concerning the parameter $\alpha_c$
 , the results reveal a non-linear trend: performance first escalates and subsequently diminishes, reaching its zenith when 
$\alpha_c$ =4. At this value, metrics like M-F, Detect, and C-F all register at their optimum, underscoring the effectiveness of the chosen approach. Such a trend suggests an inherent trade-off between diversity and accuracy. Pushing 
$\alpha_c$  towards its peak value ensures that the model does not easily overlook pivotal points in the graph, striking a balance between breadth and precision of detection.
The specific experimental data are shown in the Figure~\ref{fig:cfg}.

\paragraph{Serialization Order}
In the ablation study detailed in Table~\ref{tab:order}, conducted on the nuScenes dataset using VoVNetV2 as the image backbone, the depth-first traversal (DFS) method stands out as the most effective serialization strategy. Conversely, the $Coord_{xy}$ method, which sorts vertices based on ascending xy coordinate values, records the weakest performance. This data underscores the importance of traversal-based serialization, like DFS, in capturing spatial relationships between vertices, a crucial factor for superior lane graph detection.

\paragraph{Transformer Layers}
In Table~\ref{tab:layers}, an ablation study focusing on the number of Transformer layers is conducted on the nuScenes dataset. Among the evaluated configurations, the model with 8 Transformer layers achieves the highest scores, suggesting it is the optimal choice. The consistent increase in performance with the rising number of layers shows the importance of the large language model(LLM) in the task. The progressive enhancement with additional layers evidences the crucial role of depth in capturing intricate patterns and relationships for lane graph extraction.

\subsection{Qualitative Results}

In Figure~\ref{fig:nus_vis}, we present visualizations based on the nuScenes dataset, while Figure~\ref{fig:av2_vis} displays visualizations from the Argoverse2 dataset. 
The depicted outcomes exhibit a remarkable resemblance between our predictions and the ground truth of the overall road layout, underscoring the efficacy of our method in conducting lane graph extraction.

From Figure~\ref{fig:av2_vis}, we can find that STSU~\cite{can2021structured} exhibits inaccuracies in predicting the road layout, and TopoNet~\cite{li2023topology} displays minor deficiencies in the vicinity of junction points. Our approach effectively predicts the overall layout of the road structure with continuous and clear junction points.

\section{Conclusions}
In this work, we presented \textbf{LaneGraph2Seq}, an advanced framework that harnesses the capabilities of Transformer-based language models for the task of lane graph extraction. 
By adopting a novel vertex-edge encoding mechanism coupled with a depth-first traversal and an edge-based partition sequence, our approach adeptly captures the intricate topologies and geometries inherent in lane graphs. 
Further refinement during the inference stage, through classifier-free guidance combined with nucleus sampling, elevates the accuracy of predictions while minimizing edge false-negative rates. 
Our comprehensive evaluations on the nuScenes and Argoverse 2 datasets underscore the superiority of our method. 
Looking ahead, 
leveraging large-scale pretraining from the HD map for the large language model offers a promising direction for enhancing the robustness and generalizability of our model in diverse and challenging road scenarios.

\section{Acknowledgments}
This work was supported in part by STI2030-Major Projects (Grant No. 2021ZD0200204), National Natural Science Foundation of China (Grant No. 62106050 and 62376060),
Natural Science Foundation of Shanghai (Grant No. 22ZR1407500) and 
USyd-Fudan BISA Flagship Research Program.

\bibliography{aaai24}

\end{document}